\documentclass[runningheads]{llncs}

\usepackage{eccv}

\usepackage{eccvabbrv}

\usepackage{graphicx}

\usepackage{soul}
\usepackage{pifont}
\usepackage{tikz}
\usepackage{tikz-3dplot}
\usepackage{bm}
\usepackage{float}
\usepackage[dvipsnames,svgnames]{xcolor}
\usepackage[accsupp]{axessibility}
\usepackage[normalem]{ulem}

\usetikzlibrary{positioning, shapes.geometric}
\usetikzlibrary{calc}
\usepackage{booktabs}   %
\usepackage[table]{xcolor}   %
\usepackage{multirow}
\usepackage{placeins} %
\usepackage{wrapfig}

\usepackage{soul}
\definecolor{junglegreen}{rgb}{0.113, 0.639, 0.5}

\DeclareMathOperator*{\argmin}{arg\,min}

\usepackage{etoolbox}
\preto{\section}{\vspace{-3pt}}
\preto{\subsection}{\vspace{-2pt}}

\definecolor{RDcolor}{rgb}{0.5, 0.1, 0.8}

\newcommand{\cmark}{\textcolor{PineGreen}{\ding{51}}}%
\newcommand{\xmark}{\textcolor{BrickRed}{\ding{55}}}%

\definecolor{pastelorange}{rgb}{1.0, 0.7, 0.4}
\setulcolor{pastelorange}
\setstcolor{pastelorange}

\usepackage[accsupp]{axessibility}

\usepackage{hyperref}

\usepackage{orcidlink}

\begin{document}

\title{Towards Real-World Wearable Motion Reconstruction}

\titlerunning{Towards Real-World Wearable Motion Reconstruction}

\author{Andrea Boscolo Camiletto\inst{1,2} \and
Rishabh Dabral\inst{1,2} \and
Eduardo Alvarado\inst{1} \and
Thabo Beeler\inst{3} \and
Marc Habermann\inst{1,2} \and
Christian Theobalt\inst{1,2}}

\authorrunning{A.~Boscolo Camiletto et al.}

\institute{Max Planck Institute for Informatics, Saarland Informatics Campus, Germany \and
Saarbrücken Research Center for Visual Computing, Interaction and AI, Germany \and
Google, Switzerland}

\maketitle

\begin{figure}[H]
	\centering
	\includegraphics[width=0.95\linewidth]{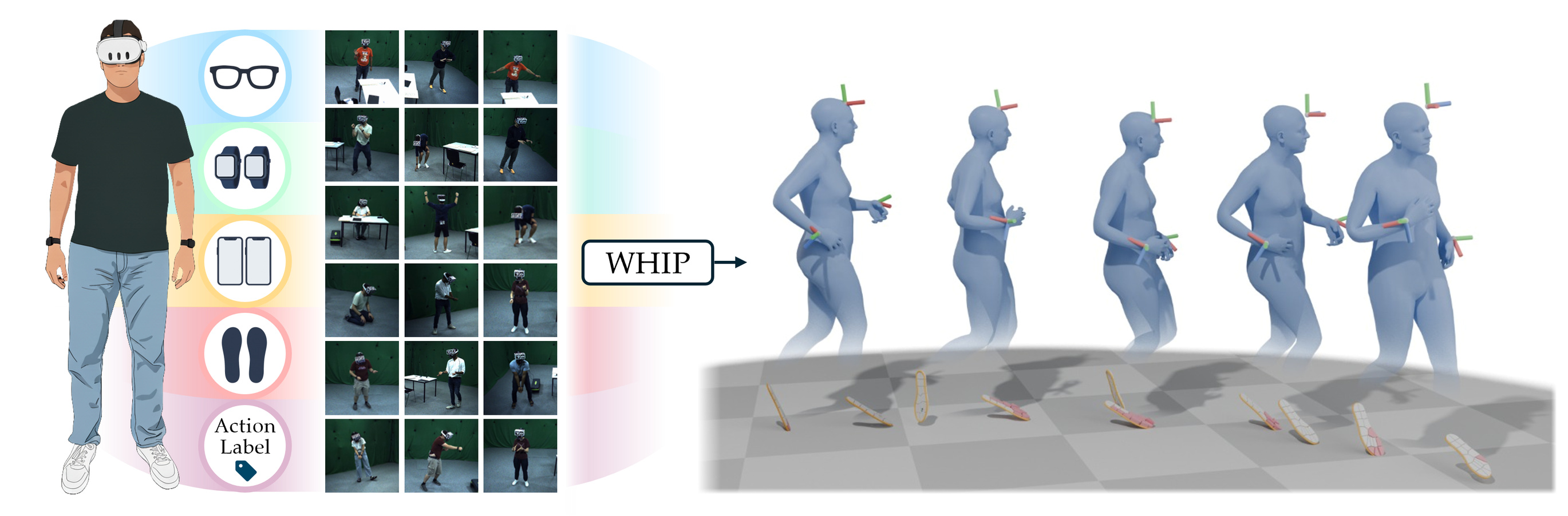}
	\caption{We present a multimodal wearable sensor suite for motion capture (left) and show full-body motion reconstructed by our model from sparse sensor readings (right).}
	\label{fig:teaser}
\end{figure}

\begin{abstract}
The modern-day surge in popularity of wearable devices poses a fundamentally unique motion capture problem: reconstructing full-body movement from any set of sensing hardware worn at a given moment.
Yet, most research efforts assume fixed sensor configurations (e.g., IMU suits or HMD-centric rigs) and cannot generalize across them.
In contrast, we argue that motion capture should prioritize unobtrusive and lightweight devices such as smartphones, smartwatches, smart glasses, and smart insoles, and study the interplay between them.
To this end, we make three contributions.
First, we present a large-scale multimodal dataset synchronizing these consumer-grade sensors with ground-truth 3D motion, spanning 50 diverse activities including everyday tasks, sports, and social interactions.
Second, we propose WHIP, a baseline generative model that reconstructs motion from arbitrary subsets of available sensors, robustly handling missing modalities and producing physically plausible motions.
Third, we conduct a systematic study of sensor complementarity, quantifying how different modalities complement one another.
Code and dataset are available at \href{https://vcai.mpi-inf.mpg.de/projects/WHIP}{this URL}.
\end{abstract}

\section{Introduction} \label{sec:intro}

Human Motion Capture (MoCap) is essential across numerous fields, including gaming, sports, medicine, VR/AR, robotics, and filmmaking, enabling the accurate tracking and reproduction of human movements.
Traditional MoCap solutions, however, present challenges for real-world deployment: optical marker-based systems \cite{vicon} achieve excellent precision but require cumbersome suits and controlled environments, whereas markerless vision-based systems require extensive camera infrastructure \cite{captury} for high-quality results.
This has motivated a shift toward lightweight, accessible sensing: smartphones and smartwatches are now ubiquitous, and smart insoles and head-mounted devices increasingly common.
Together they offer complementary cues that no single modality provides, making them a promising basis for untethered motion capture.
\par 
However, converting such sparse, commodity signals into reliable full-body tracking remains a challenging and underconstrained problem.
Prior work has addressed individual modalities in isolation. IMU-based methods use as few as six sensors \cite{dip, sip, pip, transpose}, yet even these multi-sensor setups require careful calibration and remain impractical for daily use.
A parallel line of work leverages VR/AR hardware: headsets and controllers provide native head and hand tracking, often achieving strong reconstruction accuracy \cite{avatarposer, questsim}. However, inferring leg motion from only upper-body cues is fundamentally ambiguous. 
Headsets with body-facing cameras partially mitigate this ambiguity by directly observing the body \cite{egoglass, unrealego, boscolo2025frame}, yet occlusions and limited hardware availability still limit their practicality.  
A complementary direction of increasing interest is plantar pressure sensing: by measuring foot–ground interactions, it provides informative cues on gait, balance, and contact events \cite{footplantar}. This trend is fueled by the growing availability of commercial smart insoles \cite{moticon_opengo, novel_pedar, novel_loadsol, nurvv_run}, enabling practical out-of-lab use. These signals improve motion plausibility when fused with other inputs \cite{mmvp, diffusionposer} and can even support standalone capture \cite{soleposer, dualmode_insoles}, though datasets remain scarce and many systems still require controlled settings.

In practice, people wear diverse combinations of devices, and wearable motion capture should adapt to whichever subset is available. Existing methods, however, are specialized to fixed hardware configurations.
To bridge this gap, we introduce a large-scale dataset that synchronizes everyday consumer wearables (two smartphones, two smartwatches, pressure-sensing insoles, and a VR headset) with high-quality 3D motion from a calibrated 120-camera markerless system \cite{captury}, time-aligned and spatially registered across modalities. The dataset spans 14 participants (two sessions each), 50 action classes, and over 7 hours ($\approx800$k frames) of synchronized recordings, captured using commodity hardware in realistic configurations (see Fig.~\ref{fig:teaser}).
Leveraging this data, we develop WHIP, a flow-matching generative model that reconstructs full-body motion from arbitrary sensor subsets. 
WHIP builds on a DiT-style backbone with per-modality cross-attention, conditioned on any available sensors and optional action labels, and is trained with a conditional flow-matching objective to produce temporally consistent motions.
Together, dataset and model enable a principled analysis of sensor complementarity: we systematically evaluate single- and multi-modal subsets, quantify each sensor's contribution, and identify Pareto-optimal combinations under practical hardware constraints.
\par 
In summary, our work presents:
\begin{itemize}
    \item A large-scale dataset combining synchronized everyday wearables.  
    \item WHIP, a flow-matching model that reconstructs motion from any sensors.
    \item The first cross-modal study that comprehensively compares the complementarity of individual wearable sensors for motion capture.
\end{itemize}

\section{Related Work}
\label{sec:related}

\par \textbf{Datasets for Wearable Motion Capture.}
Collecting datasets for wearable motion capture is inherently challenging, requiring precise temporal synchronization and spatial alignment with full-body tracking labels.

\emph{IMU-based datasets.}  
Several datasets have paired inertial recordings with high-quality motion capture supervision. Most notably, DIP-IMU \cite{dip} and TotalCapture \cite{totalcapture} collected data using a full-body Xsens suit, establishing standardized evaluation on a subset of sensors and the widely adopted ``six-IMU'' configuration that underpins much subsequent IMU-based pose estimation. However, they rely on intrusive full-body suits, limiting applicability in everyday scenarios. IMUPoser \cite{imuposer} reduced this overhead by relying solely on commodity devices---two smartphones, two smartwatches, and a pair of earbuds---but the resulting dataset comprises only two hours of data.

\emph{HMD-based datasets.}  
Head-mounted devices (HMDs) are another major source of wearable signals. Ego4D \cite{ego4d} and Ego-Exo4D \cite{egoexo4d} introduced large-scale egocentric video benchmarks. Nymeria \cite{nymeria} further scaled AR-glasses recordings, combining them with wristband sensors. However, their full-body tracking labels are derived from sparse external cameras, resulting in lower-quality ground truth compared to professional motion capture.

\emph{Pressure datasets.}
Research on plantar pressure and ground reaction forces has primarily targeted biomechanics rather than full-body pose. Large-scale resources pair MoCap with force plates (e.g., GroundLink \cite{groundlink}, AddBiomechanics \cite{werling2024addbiomechanics}) or pressure mats with video (MMVP \cite{mmvp}, MOYO \cite{tripathi2023ipman}), but none involve wearable insoles. Dedicated insole datasets remain small-scale: recent works coupling insoles with MoCap \cite{mourot2022underpressure,soleposer,p2pinsole} show promising pose reconstruction yet cover only narrow motion taxonomies. The complementary role of plantar signals with respect to other wearable modalities remains unexplored.

Overall, prior datasets have been siloed by modality, each advancing its own line of work without studying how these signals interact. Our dataset bridges this gap by jointly capturing smartphones, smartwatches, smart insoles, and a VR headset, all synchronized with a multi-view markerless motion capture system.

\par \noindent \textbf{Wearable Motion Capture Methods.}
\emph{IMU-based methods.}
Sparse IMU configurations have long been explored for pose reconstruction. Early optimization methods fit body models to sensor orientations and accelerations \cite{sip}, demonstrating feasibility at high computational cost. Deep learning enabled real-time solutions: DIP \cite{dip} regressed joint rotations from six IMUs using recurrent networks, inspiring later refinements with kinematic constraints \cite{transpose}, physics priors \cite{pip}, terrain awareness \cite{tip}, or part-based modeling \cite{dynaip}. These advances consolidated six-IMU setups as the research standard, with some work extending to three IMUs \cite{progip}.  
Recent approaches turned to commodity devices: IMUPoser \cite{imuposer} showed that arbitrary combinations of phones, watches, and earbuds suffice for pose estimation, while MobilePoser \cite{xu2024mobileposer} further added root translation and lightweight physics smoothing.
DiffusionPoser \cite{diffusionposer} similarly targets arbitrary sparse sensors with a generative autoregressive diffusion model, but focuses on inertial inputs rather than the broader mix of consumer wearables studied here.

\emph{HMD-based methods.}
Head-mounted devices provide another path toward minimal sensing. AvatarPoser \cite{avatarposer} and QuestSim \cite{questsim} reconstruct body motion from only head and hand poses, while EgoEgo \cite{egoego} augments HMDs with egocentric cameras to mitigate lower-body ambiguity. Diffusion models have also been adopted \cite{bodiffusion, agl} to regularize the reconstruction. 

\emph{Insole-based methods.}
Plantar pressure signals have recently gained attention. SolePoser \cite{soleposer} and P2P-Insole \cite{p2pinsole} combined pressure distributions with foot-mounted IMUs to estimate motion from sparse inputs. However, these methods have only been validated on narrow locomotion tasks.

Despite this progress, each method specializes in a single modality and a fixed sensor set. Real-world use, however, is inherently cross-modal, with sensors appearing in diverse and unpredictable combinations.
In the next section, we present the dataset we collected to study this scenario.

\section{Dataset}
\label{sec:dataset}

In this work, we aim for a real-world motion tracking model that operates with unobtrusive devices and adapts to whatever sensors are available.
\begin{wrapfigure}[15]{r}{0.55\columnwidth}
    \centering
    \vspace{-5mm}
    \includegraphics[width=0.53\columnwidth]{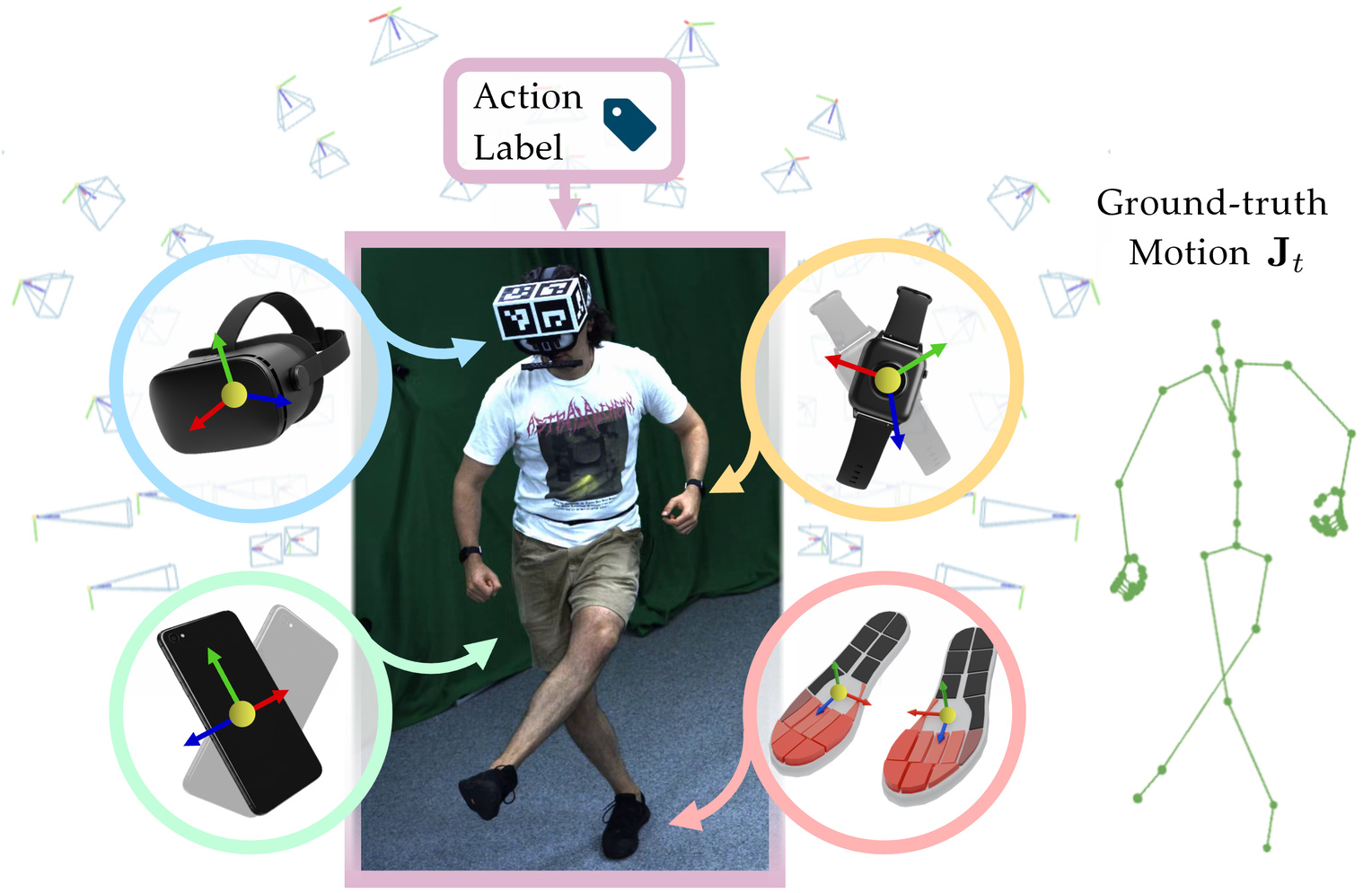}
    \caption{\textbf{Wearable capture rig.} Two smart insoles (pressure\,+\,IMU), two smartwatches and two smartphones (IMUs), and an HMD (6-DoF pose), all synchronized with markerless MoCap. }
    \label{fig:setup}
    \vspace{-9mm}
\end{wrapfigure}
This requires properties that existing datasets lack: \emph{sparse, low-overhead} and \emph{diverse sensing}.

To address this gap, we collected a multimodal dataset in which participants wore two smartphones, two smartwatches, a head-mounted device, and pressure sensing insoles, synchronized with professional markerless motion capture (see Fig.~\ref{fig:setup} and Fig.~\ref{fig:teaser}, left). All devices are commodity hardware, reflecting configurations possible in everyday use.

\subsection{Capture System}
\label{subsec:capture}
Our data acquisition takes place in a markerless motion-capture studio equipped with $120$ synchronized and calibrated RGB cameras operating at $30\,$Hz \cite{captury}.
This system provides ground-truth 3D skeletal motion, 
$\mathbf{J}_t \in \mathbb{R}^{J \times 3}$,
where $t$ denotes the time index, $J$ the number of skeletal joints and $3$ indicates the 3D coordinates in the studio reference frame.

\par
\noindent \textbf{Smartphones and Watches.}  
Participants carry two smartphones in the left and right front trouser pockets and wear a smartwatch on each wrist. Each device records an orientation estimate $\mathbf{R}_{D,t}^{S} \in \mathrm{SO}(3)$, angular velocity $\boldsymbol{\omega}_{D,t}^{S} \in \mathbb{R}^3$, and linear acceleration $\mathbf{a}_{D,t}^{S} \in \mathbb{R}^3$, all in the local device frame, where $D \in \{P,W\}$ denotes the device type and $S \in \{L,R\}$ the body side.
While these devices can expose data in a shared global frame by leveraging a magnetometer, we find such estimates often unreliable and noisy, consistent with prior reports \cite{apple_coremotion,imuposer}.
We therefore operate in the device-local frames.

\par
\noindent \textbf{Headset.}
We leverage a Meta Quest~3 headset \cite{meta_quest3} to obtain 6-DoF head tracking from its onboard SLAM, streamed via a Unity application as $\mathbf{H}_t \in \mathrm{SE}(3)$.

\par
\noindent \textbf{Insoles.}
Participants also wear Moticon OpenGo pressure-sensing insoles~\cite{moticon_opengo}. The per-foot measurements are $\mathbf{I}_t^{S}=\big[\;\mathbf{p}_t^{S},\;\boldsymbol{\omega}_{I,t}^{S},\;\mathbf{a}_{I,t}^{S}\;\big] \in\mathbb{R}^{22}$
 where \(\mathbf{p}_t^{S}\in\mathbb{R}^{16}\) are pressures from 16 plantar zones,
\(\boldsymbol{\omega}_{I,t}^{S}\in\mathbb{R}^{3}\) is the angular velocity, and
\(\mathbf{a}_{I,t}^{S}\in\mathbb{R}^{3}\) the linear acceleration from the embedded IMU.
Unlike the phones and watches, the insole IMU provides raw sensor signals.

As these sensors record at different rates, we subsample all signals to $30$\,Hz and align them to the motion capture system.

\subsection{System Calibration}
Each device records in its own local coordinate system and maintains an independent internal clock. To make the signals mutually consistent, they must be spatially aligned and temporally synchronized.

To that end, at the start of every session, participants perform a one-minute calibration sequence designed to excite all sensors, from which we resolve spatial alignment and temporal synchronization.
In the following, we describe the calibration procedures for each device.

\subsubsection{Headset.}
Following Camiletto et al.~\cite{boscolo2025frame}, we attach an ArUco board to the HMD to align it with the MoCap.
By tracking the board's pose $\mathbf{N}_t \in \mathrm{SE}(3)$ from the studio cameras and collecting the HMD inside-out tracking $\mathbf{H}_t \in \mathrm{SE}(3)$, we solve
\begin{equation}
\mathbf{T}^*_c,\, \mathbf{T}^*_r,\, t^*_0 = 
\argmin_{\mathbf{T}_c,\, \mathbf{T}_r,\, t_0} 
\sum_{t=1}^{T} 
\big\lVert 
\mathbf{N}_t - \mathbf{T}_c \, \mathbf{H}_{t+t_0} \, \mathbf{T}_r
\big\rVert^2,
\end{equation}
where $\mathbf{T}_c$ and $\mathbf{T}_r$ are the coordinate and mounting transformations, and $t_0$ is the clock offset, jointly recovering spatial and temporal alignment.

\subsubsection{Insoles.}
The Moticon OpenGo insoles $\mathbf{I}_t^{S}$~\cite{moticon_opengo} are initialized using the manufacturer’s calibration. 
Absolute timestamps are obtained via the companion app’s QR-code mechanism, which we use for alignment with the studio clock. 
This alignment, however, is too coarse: we observe clock drift both relative to the studio and between the two insoles.
To account for this, we ask participants to perform a jumping sequence towards the end of the recording, which allows us to manually adjust offsets and correct for both absolute and relative drift.

\subsubsection{Phones and Watches.}
We align each phone/watch orientation to the MoCap anatomical joint rotation (hip for phones, wrist for watches). For each device type $D\!\in\!\{P,W\}$ and side $S\!\in\!\{L,R\}$, we estimate: (i) a coordinate mapping $\mathbf{R}_c^{D,S}\!\in\!\mathrm{SO}(3)$ from the device-local frame to the studio frame, (ii) a mounting offset $\mathbf{R}_r^{D,S}\!\in\!\mathrm{SO}(3)$ between device and joint, and (iii) a clock offset $t_0^{D,S}$ between the device and ground truth.
We do so by solving the following optimization, omitting the superscripts for clarity:
\begin{equation}
\min_{\mathbf{R}_c,\,\mathbf{R}_r,\,t_0}
\sum_{t=1}^{T}
d_{\mathrm{SO}(3)}\!\left(
\mathbf{R}_c\, \mathbf{R}_{t+t_0}\, \mathbf{R}_r,\;
\mathbf{M}_{t}
\right)^{2},
\label{eq:dev-calib}
\end{equation}
where $\mathbf{R}_{t} \in \mathrm{SO}(3)$ denotes the device-reported orientation, $\mathbf{M}_{t} \in \mathrm{SO}(3)$ is the corresponding MoCap joint rotation, and $d_{\mathrm{SO}(3)}$ is the geodesic distance on $\mathrm{SO}(3)$. The same problem is solved independently for each device/side pair.

To compensate for sensor drift and for shifts in how sensors are worn, we recompute $\mathbf{R}_r^{D,S}$ for each $20$s action block. While this is an idealized setting, it compensates for misalignment that would accumulate over long sessions. In practice, the magnitude of this misalignment depends on both the sensor and how it is worn: wrist-strapped devices remain comparatively stable, whereas phones can shift inside pockets and introduce larger orientation biases. Dynamic on-body calibration methods, such as Transformer IMU Calibrator~\cite{zuo2025transformer}, are a promising way to relax this assumption in future deployments.

\subsection{Dataset Structure}
Our dataset includes 14 participants, each contributing two sessions to increase variability in sensor placement. Every session begins with a one-minute calibration, followed by 50 predefined actions (20\,s each, separated by 4\,s pauses) spanning daily and sports-related movements. The headset is operated in passthrough mode so that participants can naturally interact with their environment.

All modalities are time-synchronized at discrete steps. %
Each 20\,s segment is annotated with the respective action label.
The resulting dataset comprises over 7 hours of multimodal recordings ($\approx 800\text{k}$ synchronized frames). 
To assess generalization, we hold out one participant and three actions for evaluation.

\begin{table}[tb]
\centering
\caption{Comparison of wearable MoCap datasets. HQ-GT: high-quality ground truth from multi-camera or marker-based systems. cIMUs: commodity IMUs from consumer devices, as opposed to full IMU suits.}
\resizebox{0.80\linewidth}{!}{%
\begin{tabular}{lcccccccc} 
\toprule
Dataset                         & Insoles  & HMD    & Action  & HQ-GT & cIMUs    & Subjects & Hours \\ 
\midrule
TotalCapture \cite{totalcapture}& \xmark & \xmark & \cmark      & \cmark & \xmark   & 5     & 9    \\
DIP-IMU \cite{dip}              & \xmark & \xmark & \xmark      & \xmark & \xmark   & 10    & 1.5  \\
VT-NMD \cite{vt_nmd}            & \xmark & \xmark & \xmark      & \xmark & \xmark   & 16    & 40   \\
CIP \cite{cip_dataset}          & \xmark & \xmark & \xmark      & \xmark & \xmark   & 31    & 16   \\
\midrule
UrbanWalk \cite{urban_walk}     & \cmark & \xmark & \cmark      & \xmark & \xmark   & 20    & 9    \\
UnderPressure \cite{mourot2022underpressure}  & \cmark & \xmark & \cmark & \xmark & \xmark & 10 & 5.6 \\
\midrule
Ego-Exo4D \cite{egoexo4d}       & \xmark & \cmark & \cmark      & \xmark & \cmark  & 740   & 221  \\
Nymeria \cite{nymeria}          & \xmark & \cmark & \cmark      & \xmark & \cmark  & 264   & 300  \\
\midrule
IMUPoser \cite{imuposer}        & \xmark & \xmark & \cmark      & \cmark & \cmark  & 10    & 1.2  \\
\textbf{Ours}                   & \cmark & \cmark & \cmark      & \cmark & \cmark  & 14    & 7    \\
\bottomrule
\end{tabular}%
}
\label{tab:datasets}
\vspace{-2mm}
\end{table}

Tab.~\ref{tab:datasets} situates our dataset among prior efforts in wearable motion capture. 
Unlike earlier benchmarks that rely on dense IMU suits, lack cross-modality coverage, or forgo high-quality ground truth, our dataset combines consumer-grade smartphones, smartwatches, insoles, and a VR headset, all synchronized with professional multi-camera MoCap, thus providing both reliable 3D labels and systematic cross-modal coverage.

\section{Method}
We now present WHIP (Watch, Headset, Insole, Phone), a generative framework for reconstructing full-body motion from arbitrary combinations of consumer wearables, which serves as our baseline for studying sensor complementarity.
Leveraging the dataset from \cref{sec:dataset}, our architecture activates a dedicated processing pathway for each available sensor modality, scaling computation with the inputs present.
The model synthesizes plausible human motions conditioned on any available sensor subset and optional action labels.

Formally, we denote a motion sequence as $X = \{ \mathbf{J}_t \}_{t=1}^T$, where $\mathbf{J}_t \in \mathbb{R}^{J \times 3}$. Here, $J$ is the number of skeletal joints and $T$ is the sequence length. We use $t \in \{1, \ldots, T\}$ for frame indices and reserve $\tau \in [0,1]$ for the flow-matching time parameter.
The conditioning variables are $\mathcal{C} = \{ \mathbf{W}, \mathbf{H}, \mathbf{I}, \mathbf{P}, \mathcal{A} \}_{\mathcal{M}}$.
Here, $\mathcal{M}$ denotes an arbitrary subset of the available inputs: smartphones ($\mathbf{P}$), smartwatches ($\mathbf{W}$), insoles ($\mathbf{I}$), headset pose ($\mathbf{H}$), and action labels ($\mathcal{A}$).
\subsection{Flow Matching Framework}
\label{subsec:flow_matching}
To capture the inherent ambiguity of mapping sparse sensor signals to human motion, we adopt a probabilistic formulation based on \emph{Flow Matching} \cite{lipman2022flow}.
Our objective is to learn a velocity field $u_\tau: [0,1] \times \mathbb{R}^D \to \mathbb{R}^D$ that transports samples from a Gaussian prior to the target motion distribution, conditioned on the available sensors.
Here $D = T \times J \times 3$ denotes the dimensionality of a motion sequence.

The velocity field is induced by the probability path $p_{\tau}$ between the prior $p_0 = \mathcal{N}(0,I)$ and the data distribution $p_1$.
The evolution of a sample $x \sim p_0$ is consequently governed by the ordinary differential equation
\begin{equation}
\frac{d}{d\tau}\psi_\tau(x) = u_\tau(\psi_\tau(x)),
\label{eq:ode}
\end{equation}
where $\psi_\tau$ denotes the flow map from $p_0$ to $p_{\tau}$.

Given sensor observations $\mathcal{C}$ and ground-truth motion $X_1 \sim p_1(\cdot|\mathcal{C})$, we construct a linear interpolation path inducing the conditional velocity field
\begin{equation}
u_\tau(X_\tau \mid X_1, \mathcal{C}) = \frac{X_1 - X_\tau}{1-\tau}.
\end{equation}

\begin{figure}[t]
    \centering
    \includegraphics[width=\linewidth]{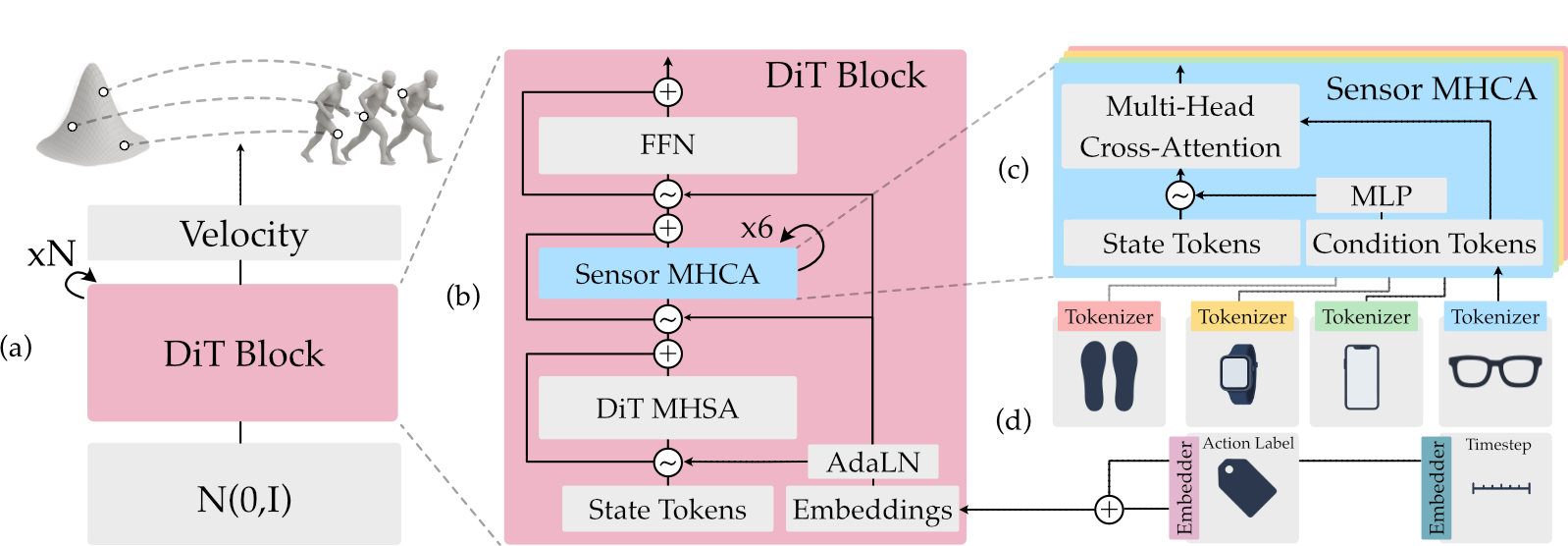}
    \caption{WHIP model architecture. (a) High-level overview, from noise to the velocity field. (b) DiT transformer block. (c) Sensor-specific cross-attention modules. (d) Conditioning information.}
    \label{fig:architecture}
\end{figure}

\subsection{Model Architecture}
WHIP builds upon the Diffusion Transformer (DiT) architecture~\cite{dit} with modifications for conditioning on heterogeneous sensors. Fig.~\ref{fig:architecture} illustrates the architecture.

\subsubsection{Conditioning Signals.}

Conditioning is processed through two pathways: global embeddings for sequence-level attributes and sensor tokens for time-aligned wearable signals (Fig.~\ref{fig:architecture}d).

\par
\noindent \textbf{Global embeddings} are sequence-level conditions comprising flow time $\tau \in [0,1]$ and action labels $a \in \mathcal{A}$.
The flow time $\tau$ is encoded through an MLP, while action labels are mapped to learnable embeddings, following DiT \cite{dit}.
Embeddings then modulate the transformer via adaptive layer normalization (adaLN).

\par
\noindent \textbf{Sensor tokens} encode the wearable signals independently at each time step. For each modality $m$, we form
\begin{equation}
\mathbf{C}^m = \{\mathbf{c}_t^m\}_{t=1}^T, \qquad \mathbf{c}_t^m = \mathrm{MLP}_m(\mathbf{s}_t^{m}),
\end{equation}
where $\mathbf{s}_t^{m}$ collects the modality-specific raw sensor inputs at time $t$. For modalities with multiple instances such as IMUs or insoles, the same MLP is applied independently to each device, yielding separate token sequences per instance.

\subsubsection{State Tokenization.}
We represent motion trajectories as sequences of per-frame tokens $\mathbf{X} = \{\mathbf{x}_t\}_{t=1}^T$ with $\mathbf{x}_t \in \mathbb{R}^{d}$, where $d$ is the transformer's feature dimension. To facilitate conditioning, we partition each token into motion and sensor-reserved subspaces: $\mathbf{x}_t = [\mathbf{x}^{\text{mot}}_t;\, \mathbf{x}^{\text{sens}}_t]$ with $\mathbf{x}^{\text{mot}}_t \in \mathbb{R}^{d_m}$ and $\mathbf{x}^{\text{sens}}_t \in \mathbb{R}^{d_{s}}$.
A linear map embeds the motion into the motion subspace, while the sensor channels are obtained by projecting each conditioning signal into a lower-dimensional subspace.

\subsubsection{Transformer Blocks.}
Our model backbone consists of 8 stacked DiT blocks operating in a latent dimension of $d=768$ with $12$ attention heads. Each WHIP block extends the standard DiT architecture with a multi-modal cross-attention mechanism for sensor conditioning (see Fig.~\ref{fig:architecture}c).

We instantiate a dedicated cross-attention module for each conditioning modality $\mathbf{C}^m$. These modules operate conditionally: if the corresponding sensor is present, the module runs; if absent, the module is not evaluated and contributes no update (i.e., it is excluded from the sum). The motion tokens $\mathbf{X}$ serve as queries to the conditioning sequences, with the resulting updates aggregated as:

\begin{equation}
\mathbf{X}' = \mathbf{X} + \frac{1}{\sqrt{|\mathcal{M}|}} \sum_{m \in \mathcal{M}} \text{CrossAttn}_m(\mathbf{X}, \mathbf{C}^m),
\end{equation}

where $\mathcal{M}$ denotes the set of active modalities. The $1/\sqrt{|\mathcal{M}|}$ normalization factor keeps the residual-connection variance consistent regardless of the number of active inputs.

This per-modality design makes the model robust to missing sensor inputs: during training, we randomly drop modalities per sample, and at inference, absent modalities are simply skipped.

\subsection{Training and Inference}

We optimize the model using the conditional flow matching objective \cite{lipman2022flow}: $\mathcal{L}(\theta) = \mathbb{E}_{\tau}\!\left[\|u_\tau^\theta(X_\tau, \mathcal{C}) - u_\tau(X_\tau|X_1, \mathcal{C})\|^2\right]$, where $\tau \sim \mathcal{U}[0,1]$, $X_0 \sim \mathcal{N}(0, I)$, and the target velocity is as defined in \cref{subsec:flow_matching}.
During inference we integrate the ODE in \cref{eq:ode} with the learned velocity field from $X_0 \sim \mathcal{N}(0, I)$ over $\tau \in [0,1]$, using an explicit Euler solver with $N$ discretization steps.

\section{Experiments}
\par \noindent \textbf{Implementation Details.}
We process motion sequences in sliding windows of $T=90$ frames (3 seconds at 30 fps) using a batch size of 64. 
The WHIP transformer comprises 8 blocks with latent dimension $d=768$ and 12 attention heads, each augmented with 6 dedicated cross-attention modules: HMD, insoles, left/right watches, and left/right phones.
Each conditioning signal is projected to a per-modality latent dimension of $64$. 
We train with AdamW \cite{adamw} using a learning rate of $10^{-3}$ and warmup-stable-decay (WSD) scheduling for 110k steps.
To ensure robust performance across diverse sensor configurations, we apply dropout during training, retaining each available sensor with probability $0.5$. 
This sampling scheme exposes the model to heterogeneous subsets while maintaining a consistent marginal frequency for every modality, which we find to be a good balance between sparse single-sensor cases and fully instrumented setups.
\par \noindent \textbf{Metrics.} 
Following standard practice, we evaluate using: Mean Per Joint Position Error (MPJPE) after root alignment, Mean Root Error (MRE), scale-normalized MPJPE (N-MPJPE), and Procrustes-aligned MPJPE (PA-MPJPE).
\par \noindent \textbf{Dataset and Evaluation.}
We evaluate generalization along two axes by withholding one subject and three action categories from the training set. 
Unlike deterministic regression methods, our generative model learns a conditional distribution $p_\theta(X|\mathcal{C})$ over plausible motions. 
While it is common to report best-of-$K$ sample performance for such models, this would unfairly favor generative methods over regression baselines. We instead evaluate by averaging multiple predictions, effectively using the Bayes estimator under $L_2$ loss:
\begin{equation}
    \hat{X} = \mathbb{E}_{X}[X] \approx \frac{1}{K}\sum_{k=1}^K X_k, \quad X_k \sim p_\theta(X|\mathcal{C}),
\end{equation}
where we empirically approximate the posterior mean using $K=10$ samples.
\begin{table}[t]
\centering
\newcolumntype{G}{>{\columncolor{gray!0}}c}
\newcolumntype{H}{>{\columncolor{gray!0}}c}
\newcolumntype{I}{>{\columncolor{gray!0}}c}
\caption{Comparison of regression baselines and our generative model across different sensor subsets.
\textbf{All} uses the full set of sensors, \textbf{HMD} the headset only, \textbf{Ins} the insoles only, \textbf{W+P} the average over all watch--phone combinations, and the \textbf{Avg} column reports averaged across all possible sensor configurations.}
\label{tab:gen_vs_reg}
\resizebox{0.90\linewidth}{!}{%
\begin{tabular}{c l c c c H I c c c}
\toprule
 & \multirow{2}{*}{Method} & \multirow{2}{*}{Params}
   & \multicolumn{2}{c}{MPJPE $\downarrow$}
   & \multirow{2}{*}{\cellcolor{gray!0} MRE $\downarrow$}
   & \multirow{2}{*}{\cellcolor{gray!0} PA-MPJPE $\downarrow$}
   & \multicolumn{3}{c}{N-MPJPE $\downarrow$} \\
\cmidrule(lr){4-5}\cmidrule(lr){8-10}
 &  &  & All & HMD &  &  & Ins & W+P & Avg \\
\midrule
\multirow{3}{*}{\shortstack{Unseen\\Actor}}
 & MLP         & 54.6M & 127.8         & 147.5          & 203.2         & 95.0          & 157.0          & 123.8          & 120.1 \\
 & Transformer & 57.4M & 78.3          & 135.9          & 125.9         & 61.8          & \textbf{147.6} & 98.9           & 93.9 \\
 & Ours        & 59.4M & \textbf{56.7} & \textbf{107.4} & \textbf{54.6} & \textbf{46.5} & 151.7          & \textbf{83.0}  & \textbf{77.9} \\
\midrule
\multirow{3}{*}{\shortstack{Unseen\\Action}}
 & MLP         & 54.6M & 128.5         & 140.9          & 146.5         & 106.6         & 142.3          & 128.7          & 125.1 \\
 & Transformer & 57.4M & 83.3          & 140.5          & 80.3          & 74.1          & 130.6          & 115.8          & 106.0 \\
 & Ours        & 59.4M & \textbf{60.0} & \textbf{106.6} & \textbf{37.5} & \textbf{55.7} & \textbf{121.8} & \textbf{99.5}  & \textbf{87.0} \\
\bottomrule
\end{tabular}%
}
\end{table}

\subsection{Generative vs. Regressive Baselines}
We first compare our generative model against regression-based baselines that directly map sensor inputs to body poses. 
For a fair comparison, all methods share the same embedding networks, and we match the number of learnable parameters across models. 
The baselines operate on flattened latent vectors obtained from the modality-specific encoders. 
When a sensor is unavailable, its corresponding latent vector is masked to zero. 
We evaluate against two standard regression approaches: an MLP and a transformer encoder, both matched in parameter count to WHIP.
Results in Tab.~\ref{tab:gen_vs_reg} show that our generative approach consistently outperforms both regression baselines across different sensor configurations, demonstrating the benefit of modeling motion as a generative process.
\subsection{Comparison to Existing Methods}
\begin{table}[t]
\centering
\caption{Comparison with prior methods, all retrained on our dataset (N-MPJPE, mm; lower is better). Wtc = smartwatch. \textbf{IMUs only} uses orientation-only IMU signals with no headset tracking: \emph{All} = all phones and watches, \emph{Head+Wtc} = headset orientation + watches, \emph{Avg} = mean over all IMU subsets. \textbf{With HMD} adds the headset's 6-DoF pose: \emph{Alone} = headset only, then \emph{+2\,Wtc} and \emph{+All} progressively add the two watches and the full sensor set.}
\label{tab:comparison}
\setlength{\tabcolsep}{3pt}
\resizebox{0.65\linewidth}{!}{%
\begin{tabular}{l c c c c c c}
\toprule
\multirow{2}{*}{Method}
  & \multicolumn{3}{c}{IMUs only}
  & \multicolumn{3}{c}{With HMD} \\
\cmidrule(lr){2-4}\cmidrule(lr){5-7}
 & All & Head+Wtc & Avg & Alone & +\,2\,Wtc & +\,All \\
\midrule
IMUPoser    & 75.5  & 91.8  & 96.9  & --    & --   & --   \\
EgoEgo      & --    & --    & --    & 108.4 & --   & --   \\
EgoAllo     & --    & --    & --    & 125.9 & --   & --   \\
MocapEvery  & --    & --    & --    & 141.2 & 86.1 & --   \\
AvatarPoser & --    & --    & --    & --    & 78.2 & --   \\
AGRoL       & --    & --    & --    & --    & 79.8 & --   \\
HMDPoser    & --    & --    & --    & --    & --   & 74.6 \\
\textbf{Ours} & \textbf{61.7} & \textbf{82.0} & \textbf{88.2} & \textbf{104.6} & \textbf{68.7} & \textbf{52.6} \\
\bottomrule
\end{tabular}
}
\end{table}

While this is the first work to address such a wide range of sensors, we can compare against existing methods on individual subsets by retraining them on our data, adapting each to our conventions and skeleton.
For more details, we refer to the supplemental material.
In particular, for IMUPoser \cite{imuposer} and HMDPoser \cite{hmdposer}, we omit IMUs on feet, pelvis, and headphones, as they are not available in our dataset.
For head-pose baselines \cite{egoego,egoallo,lee2024mocapevery}, we retrain the motion-estimation component on our data and remove visual components whenever present, retaining only the head-pose inputs available in our setting.
For 3-point tracking methods \cite{agl, avatarposer}, we zero out the controller positions since our watches provide only orientation.
Despite being trained for a more diverse task, i.e., reconstructing skeletal motion from arbitrary sensor combinations, our baseline model, WHIP, still outperforms previous works even though they have been exclusively trained for the respective set of sensors (see Tab.~\ref{tab:comparison}).
\subsection{Cross-Dataset Evaluation on Nymeria}
\begin{figure}[h]
  \centering
  \begin{minipage}[c]{0.34\linewidth}
    \centering
    \captionof{table}{WHIP on Nymeria. N-MPJPE (mm).}
    \label{tab:nymeria}
    \setlength{\tabcolsep}{3pt}
    \scriptsize
    \begin{tabular}{lccc}
    \toprule
    Input & Full & Upper & Lower \\
    \midrule
    VR & 87.8 & 85.7 & 91.2 \\
    + wrists & 76.7 & 67.3 & 91.2 \\
    + pelvis & 71.9 & 65.0 & 82.6 \\
    + contact & \textbf{68.7} & \textbf{63.8} & \textbf{76.3} \\
    \bottomrule
    \end{tabular}
  \end{minipage}
  \hfill
  \begin{minipage}[c]{0.62\linewidth}
    \centering
    \includegraphics[width=\linewidth]{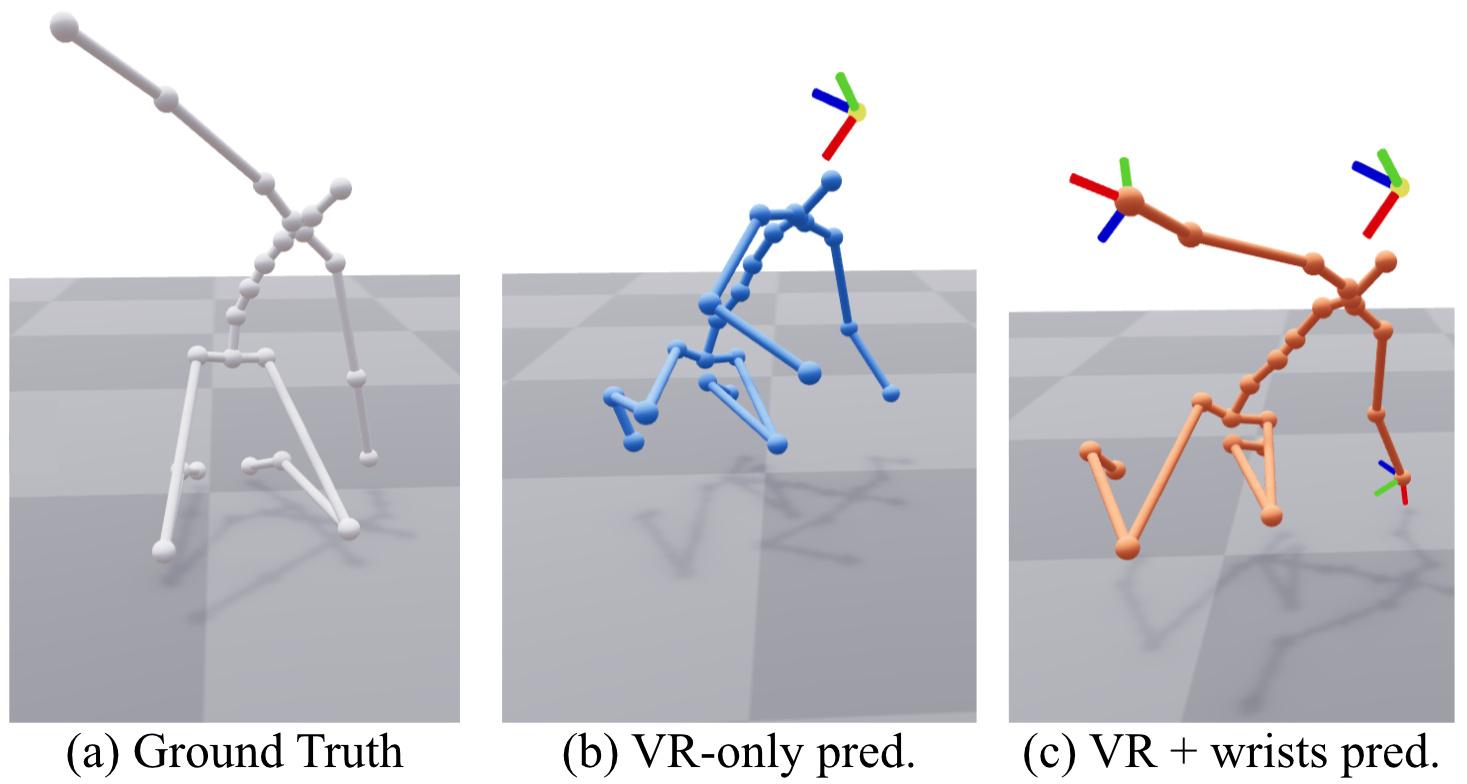}
    \caption{WHIP on Nymeria. Wrist signals recover arm motion missed by VR-only reconstruction.}
    \label{fig:nymeria_whip}
  \end{minipage}
\end{figure}

We further evaluate WHIP on Nymeria~\cite{nymeria} to test whether the model transfers to another wearable-motion dataset.
Since Nymeria does not include our exact sensor suite, we adapt the inputs by using wristband tracking in place of smartwatches, the pelvis IMU in place of phones, and labeled contacts in place of pressure insoles.
Because Nymeria has no canonical train/test split and prior work such as Ego4o~\cite{ego4o} does not release its split, our results are not directly comparable to published numbers.
Nevertheless, the trends in Tab.~\ref{tab:nymeria} are consistent with our dataset: adding wrist, pelvis, and contact cues progressively improves reconstruction, especially for the lower body.
\subsection{In-the-Wild Qualitative Results}
\begin{figure}[t]
  \centering
  \includegraphics[width=\linewidth]{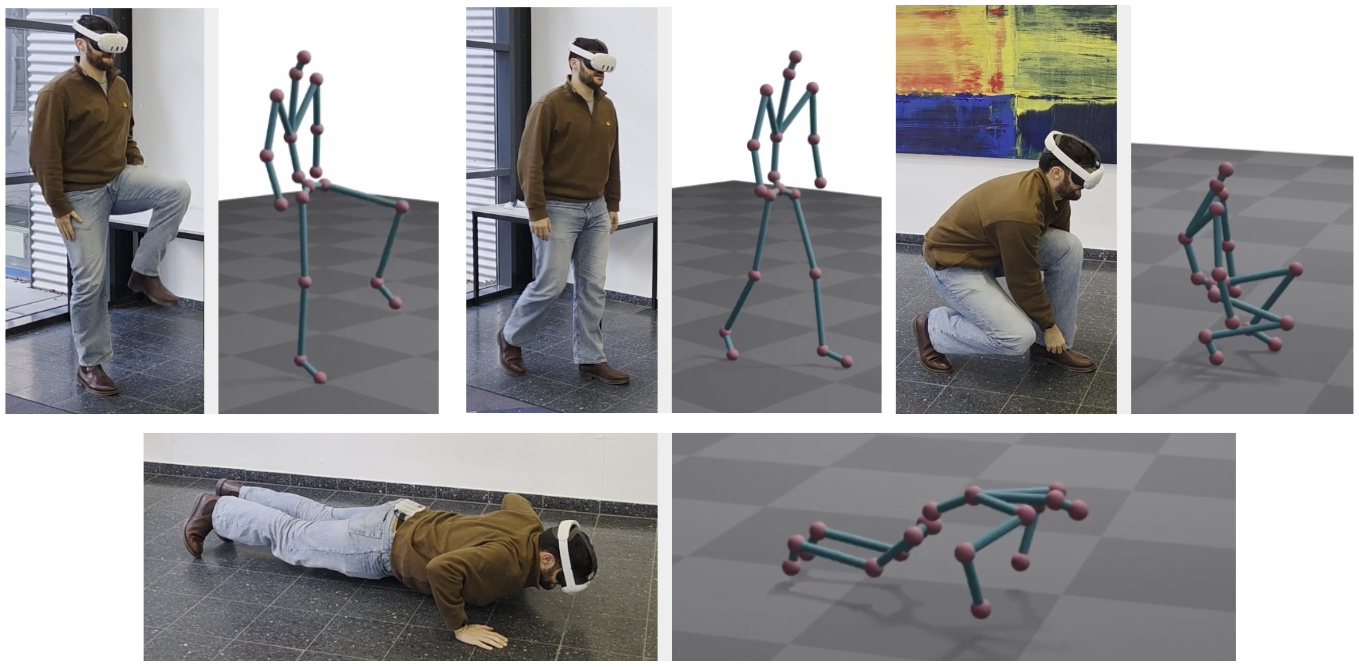}
  \caption{Qualitative results on in-the-wild recordings. Despite the domain shift, WHIP produces plausible motion reconstructions.}
  \label{fig:itw}
\end{figure}

To demonstrate applicability beyond controlled settings, we recorded additional sequences in unconstrained environments using a manual calibration procedure.
Fig.~\ref{fig:itw} shows qualitative results: despite the domain shift from our studio-captured training data, WHIP produces plausible motion reconstructions, suggesting that the learned sensor-to-motion mapping generalizes to unconstrained environments.
\subsection{Analyzing the Interplay of Modalities}
Next, we provide a statistical analysis of each individual input modality, focusing on how different sensors complement one another.
\begin{figure}[t]
  \centering
  \includegraphics[width=0.9\columnwidth]{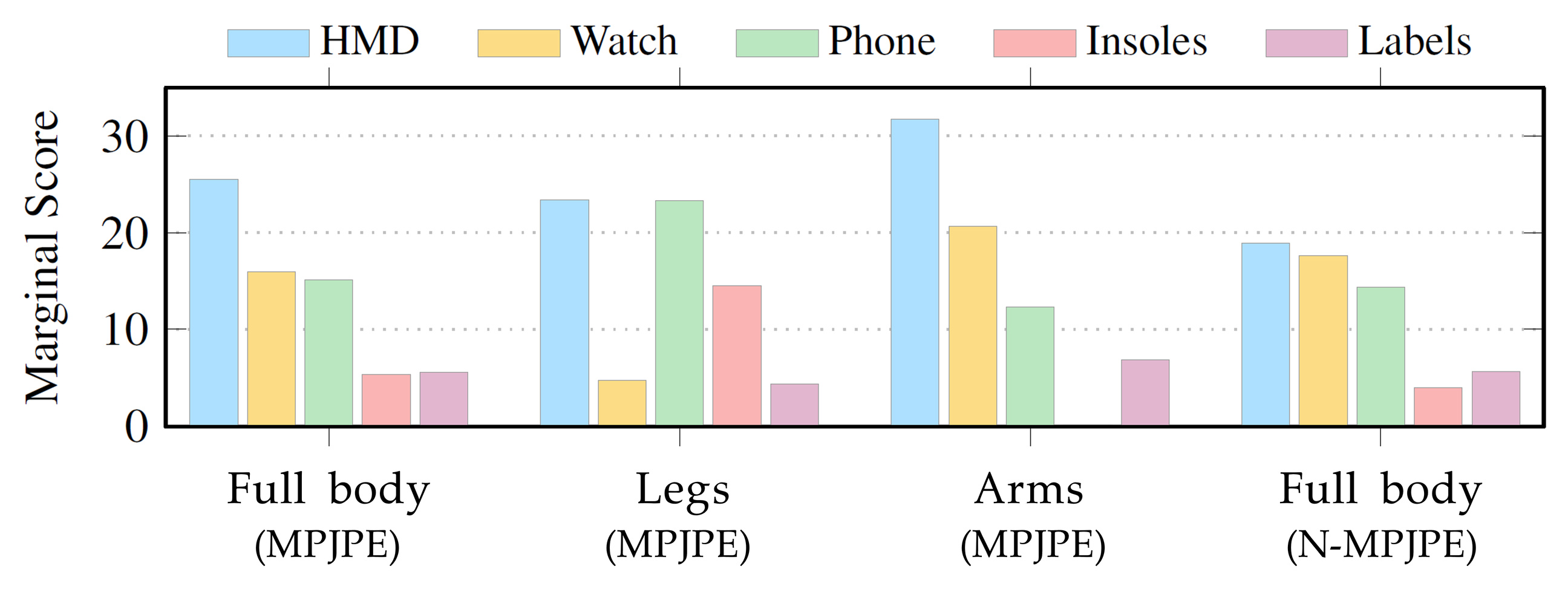}
  \caption{Sensor importance via \emph{marginal scores} across four contexts: full body, legs, arms, and normalized full body. Bars show the error reduction contributed by each modality (larger is better).}
  \label{fig:sensor_interplay_overview}
\end{figure}
\par 
For a specific set of sensors $\mathcal{C}\subseteq\mathcal{M}$ and a given metric, let $v(\mathcal{C})$ be the error (lower is better). 
With $|\mathcal{M}|=7$ inputs (HMD tracking, insoles, left/right watch, left/right phone, and action label) there are $2^{7}=128$ possible combinations; we evaluate all of them and report metrics for each.
\par \noindent \textbf{Importance of Individual Sensors.} 
To evaluate the importance of a specific modality $m\in\mathcal{M}$, we compute the expected error reduction when added to a subset $S\subseteq\mathcal{M}\setminus\{m\}$, which we refer to as the \emph{marginal score} of $m$:
\begin{equation}
\text{Marginal}(m) := \mathbb{E}_{S \subseteq \mathcal{M}\setminus\{m\}} \big[ \, v(S) - v(S\cup\{m\}) \, \big].
\end{equation}
The term $v(S) - v(S\cup\{m\})$ is the \emph{conditional} improvement of $m$ given a subset of sensors $S$. Results are reported in Fig.~\ref{fig:sensor_interplay_overview}.
Values are in millimeters for MPJPE and N-MPJPE; higher values indicate greater benefit from including that sensor.
\par 
The figure shows that (i) HMD contributes the most across all contexts, partly because it is the only sensor that recovers metric scale. When evaluated with N-MPJPE, its contribution, although still substantial, is less pronounced. 
(ii) Smartwatches are the second most important sensor overall and especially effective for arm reconstruction. 
(iii) Phones and pressure insoles primarily benefit the reconstruction of leg joints, where capturing pelvis motion and foot–ground contacts becomes central. 
(iv) Action labels provide smaller and less consistent gains than physical signals for pose accuracy.
\par \noindent \textbf{Pairwise Interactions.} 
To characterize how different sensor modalities cooperate or overlap, we quantify their \emph{pairwise interaction} $I$.
Given two sensor modalities $a,b$, and a subset of the remaining sensors $X\subseteq\mathcal{M}\setminus\{a,b\}$, we measure how their joint contribution deviates from the sum of their individual contributions:
\begin{equation}
  I(a,b)=\mathbb{E}_{X}\big[v(X_{ab})-v(X_{a})-v(X_{b})+v(X)\big]
\end{equation}
\begin{wrapfigure}{r}{0.6\columnwidth}
  \centering
  \vspace{-5mm}
  \includegraphics[width=0.58\columnwidth]{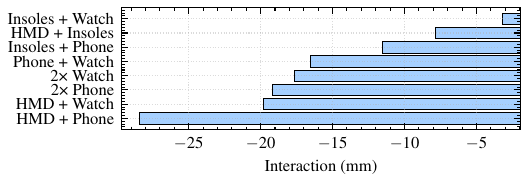}
  \caption{Pairwise sensor interactions on N-MPJPE. Higher values indicate greater synergy.}
  \label{fig:sensor_interplay_interactions}
  \vspace{-5mm}
\end{wrapfigure}
where $X_a:=X\cup\{a\}$, $X_b:=X\cup\{b\}$, and $X_{ab}:=X\cup\{a,b\}$. Since adding a sensor can only reduce (or maintain) the error, $I(a,b)\leq 0$: it is zero when the two sensors are perfectly complementary, i.e., their joint gain equals the sum of their individual gains, and increasingly negative as their contributions overlap (redundancy).
However, since $I(a,b)$ is averaged over all sensor subsets, the marginal gains of adding sensors inevitably diminish, and perfect complementarity is unattainable.
Fig.~\ref{fig:sensor_interplay_interactions} summarizes the results on N-MPJPE.
We observe that, while the insoles contribute modestly when considered alone, they exhibit the strongest synergy when paired with other modalities, providing information most distinct from other sensors.
\par
\begin{figure}[t]
  \centering
  \includegraphics[width=0.75\columnwidth]{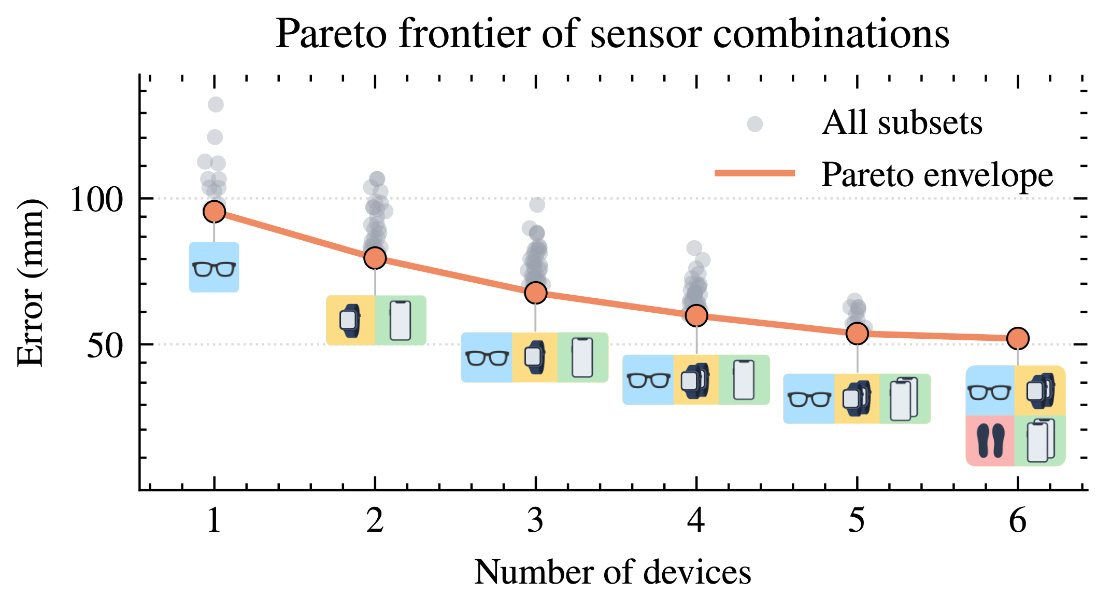}
  \caption{Pareto frontier of N-MPJPE versus number of devices. Each point represents a sensor combination; the frontier highlights the best subset for each device count.}
  \label{fig:pareto_frontier}
\end{figure}

 \noindent \textbf{Best-$k$-out-of-all Sensors.}
To quantify the trade-off between accuracy and instrumentation, we compute a Pareto frontier of reconstruction performance over sensor combinations.
For each device count $k$, we identify the subset $\mathcal{C}$ with the lowest error in terms of N-MPJPE, yielding a discrete frontier.
Fig.~\ref{fig:pareto_frontier} shows all subsets as a scatter, with the Pareto envelope highlighted.

We observe:
(i) with a single device, \emph{HMD} dominates the frontier;
(ii) among two devices, \emph{Watch+Phone} outperforms pairings that include \emph{HMD}, suggesting complementary cues across upper/lower body;
(iii) adding \emph{HMD} as a third device produces a marked gain;
(iv) additional devices yield diminishing returns, with \emph{Insoles} providing a modest improvement at full instrumentation.

\section{Limitations and Future Work}
Despite the breadth of our multimodal dataset, we did not record additional VR-centric signals such as hand-tracking from controllers, which could further improve body- and hand-focused reconstruction.
We plan to enrich future recordings with additional sensor modalities.
Our sensor-complementarity analysis uncovers consistent trends (e.g., the HMD's strong single-device performance, and the benefit of pairing insoles with smartwatches), but different conclusions may emerge when focusing on specific scenarios or activity types, where the relative importance of individual sensors could shift.
Finally, WHIP was developed and evaluated for robustness and accuracy rather than low-latency on-device operation, leaving real-time and low latency deployment as future work.
\section{Conclusion}
In this work, we introduce a large, task-diverse multimodal dataset of consumer-grade wearables synchronized with ground-truth 3D motion, surpassing existing datasets in sensor diversity and scale.
Our dataset enables unobtrusive, real-world configurations and supports a systematic analysis of sensor complementarity, quantifying how modalities interact and reinforce one another. Together, these contributions provide a valuable resource for consumer-oriented motion reconstruction methods.
In addition, our proposed generative method, WHIP, demonstrates reliable reconstruction from sparse signals and consistent gains from complementary device combinations while remaining robust to missing modalities.
Together, the dataset and model advance consumer-grade motion capture and provide a practical foundation for future work on pose reconstruction and broader in-the-wild generalization.

\paragraph{Acknowledgements.}
This work was funded by the Saarbr{\"u}cken Research Center for Visual Computing and Artificial Intelligence (VIA).

\bibliographystyle{splncs04}
\bibliography{references}

\end{document}